\documentclass[runningheads]{llncs}

\usepackage[mobile]{eccv}
\newcommand{\ours}{DreamEdit3D}

\usepackage{eccvabbrv}

\usepackage{graphicx}
\usepackage{booktabs}
\usepackage{array}
\usepackage{multirow}
\usepackage{amsmath}
\usepackage{amssymb}
\usepackage{tikz}
\usetikzlibrary{positioning, arrows.meta, fit, backgrounds, calc}
\usepackage{xcolor}
\usepackage{float}
\usepackage[normalem]{ulem}

\usepackage[accsupp]{axessibility}  %

\usepackage[breaklinks,colorlinks,citecolor=eccvblue]{hyperref}

\usepackage{orcidlink}

\begin{document}

\title{\ours{}: Personalization of Multi-View Diffusion Models for 3D Editing}
\author{Jinxin Ai\inst{1}\orcidlink{0009-0005-0977-2378} \and
Matthias Nießner\inst{1}\orcidlink{0000-0001-6093-5199} \and
Ziya Erko\c{c}\inst{1}\orcidlink{0000-0003-2656-3680}}

\titlerunning{\ours{}}
\authorrunning{J.~Ai et al.}
\institute{Technical University of Munich}

\maketitle

\begin{figure}[H]
  \centering  \includegraphics[width=\linewidth]{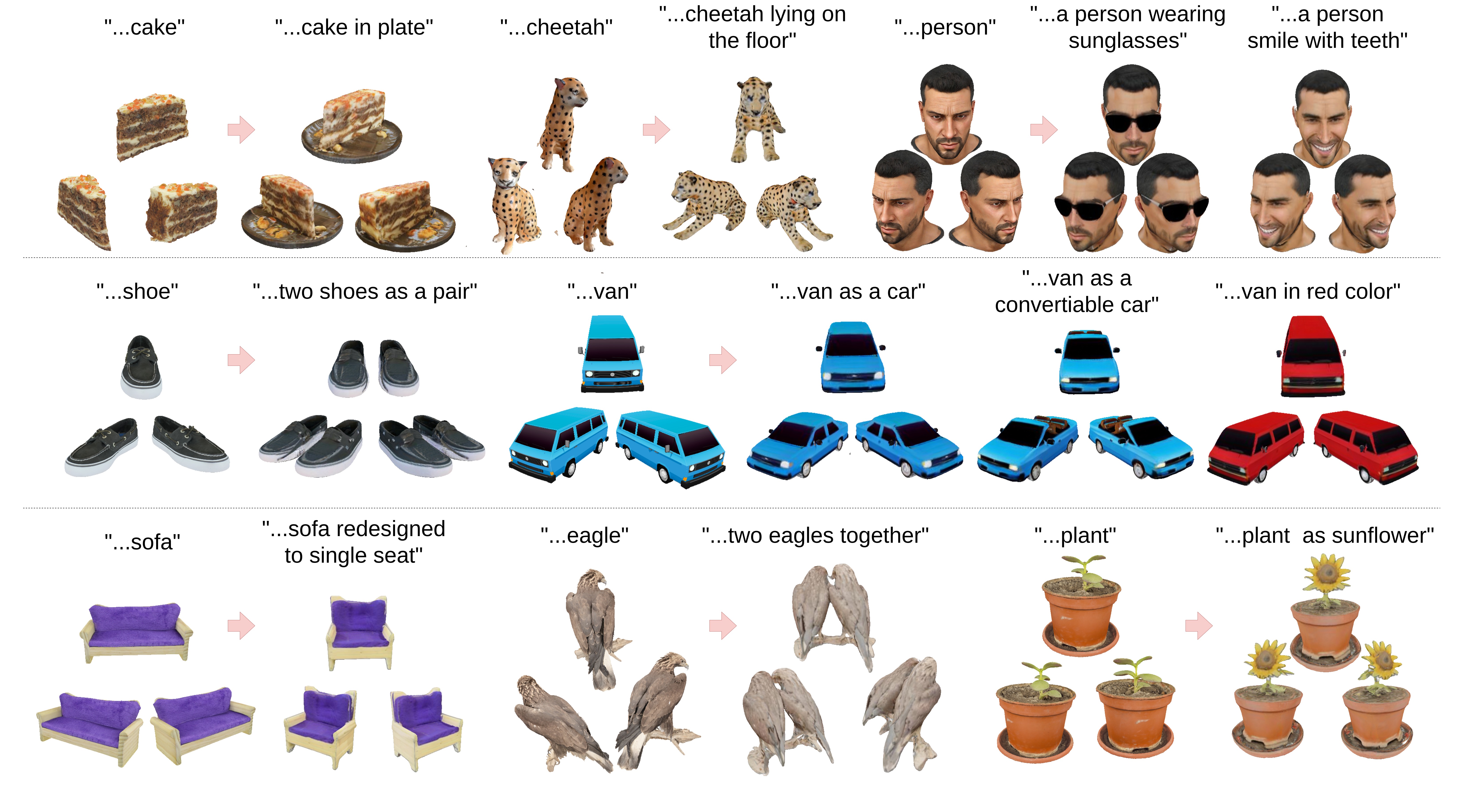}
  \caption{\textbf{\ours{}} produces multi-view consistent edits guided by natural language, given a source 3D object. We apply personalization to multi-view diffusion models to preserve the identity of the input shapes. We show that multiple diverse edits can be generated from one source by preserving the input.}
  \label{fig:teaser}
\end{figure}

\begin{abstract}
  While 2D diffusion models have achieved remarkable success in identity-preserving personalization, extending this capability to 3D assets remains a significant challenge due to the complexities of multi-view consistency and spatial control. Inspired by these 2D advancements, we present a novel personalization method for text-guided 3D editing that enables compositional, object-level control through natural language. Given a 3D input, we render orthogonal views and extract object-level segmentation masks to isolate semantic components. We then learn distinct token embeddings for each component through a tailored two-phase optimization strategy: multi-view textual inversion with attention alignment, followed by full fine-tuning of multi-view diffusion model. During inference, these disentangled tokens seamlessly compose with editing prompts to generate multi-view consistent images, which are subsequently lifted into high-fidelity textured 3D meshes. Extensive evaluations across diverse editing scenarios demonstrate that our method successfully transfers the flexibility of 2D personalization to 3D, achieving state-of-the-art edit faithfulness and identity preservation compared to existing baselines.
  \keywords{3D editing \and Diffusion models \and Textual inversion \and Multi-view generation}
\end{abstract}

\section{Introduction}
The rapid evolution of diffusion models has revolutionized the generation of visual content, extending remarkably from 2D images to 3D assets. Recent text-to-3D methodologies \cite{poole2022dreamfusion, lin2023magic3d,wang2023prolificdreamer} have demonstrated impressive capabilities in synthesizing novel 3D objects from natural language descriptions. However, the editing of existing 3D assets remains a formidable challenge. In practical modeling workflows, users rarely want to regenerate an entire asset from scratch; rather, they require precise, compositional control to modify specific semantic parts of an object while strictly preserving the identity and geometry of the unedited regions. Existing text-guided 3D editing methods~\cite{erkocc2025preditor3d,mvedit2024,sella2023vox} often struggle with this, as global text prompts tend to entangle attributes, leading to unintended global modifications that destroy the original asset's core identity.

A compelling solution to identity preservation has recently emerged in the 2D domain. Personalization techniques, such as Textual Inversion and DreamBooth, have achieved remarkable success by learning unique token embeddings for specific subjects, allowing them to be seamlessly synthesized in novel contexts, styles, and forms without losing their defining characteristics~\cite{gal2022image, ruiz2023dreambooth, avrahami2023bas}. Naturally, extending this paradigm to 3D editing is highly desirable. Yet, directly lifting 2D personalization to 3D is non-trivial. It introduces two major bottlenecks: the necessity of maintaining strict multi-view consistency across generated views, and the difficulty of spatially disentangling complex 3D objects into independently controllable semantic parts within the diffusion latent space.

Inspired by the success of 2D identity preservation, we present a novel, disentangled personalization framework designed specifically for compositional, text-guided 3D editing. Our core insight is that by explicitly isolating semantic components in the multi-view image space, we can learn distinct, disentangled token embeddings that bring the precise control of 2D personalization into the 3D domain.

To achieve this, our framework begins by rendering four orthogonal views of a given 3D input and extracting object-level segmentation masks to isolate its semantic parts. To ensure each learned token accurately represents its corresponding 3D component without bleeding into others, we propose a tailored, two-phase optimization strategy. In the first phase, we perform multi-view textual inversion guided by an attention alignment mechanism, forcing the cross-attention maps of the learned tokens to align precisely with the extracted segmentation masks across all views. In the second phase, we perform full UNet fine-tuning via joint multi-view training, which embeds strong multi-view consistency and structural priors directly into the model.

At inference, this disentangled representation unlocks highly flexible, object-level control. The optimized tokens can be combined with natural language editing prompts to independently modify specific parts of the subject. A multi-view diffusion model then synthesizes consistent, edited images across all views, which are finally lifted back into a high-fidelity, textured 3D mesh using 3D reconstruction techniques \cite{hong2023lrm}. Extensive experiments demonstrate that our approach successfully translates the power of 2D personalization to 3D, allowing for complex, localized edits that maintain the highest fidelity to the original asset's unedited regions.

\begin{itemize}
  \item We propose a disentangled token learning for multi-view diffusion models that decomposes 3D objects into editable semantic components through personalized token embeddings.
  \item We demonstrate the effectiveness of our approach on a diverse benchmark of editing scenarios, achieving favorable results compared to state-of-the-art on widely-used metrics such as CLIP and VLM-based evaluation.
\end{itemize}

\section{Related Work}
\label{sec:related}

\subsection{2D Editing}
Early image editing methods operated in the latent spaces of GANs \cite{goodfellow2020generative}, enabling semantic manipulation through latent direction traversal \cite{karras2019style}, but were limited by inversion fidelity and the expressiveness of the learned space.
The rise of text-to-image diffusion models \cite{rombach2022latent,ramesh2022hierarchical,baldridge2024imagen,BetkerImprovingIG,esser2024scalingrectifiedflowtransformers} shifted editing toward prompt-driven manipulation via vision-language alignment. To improve spatial precision, Prompt-to-Prompt \cite{hertz2022prompt} redirects edits through cross-attention map manipulation, while inpainting-based methods \cite{lugmayr2022repaint} allow region-specific modification. There have been significant improvements in personalization-based editings to preserve the identity of the input~\cite{ye2023ip,ruiz2024hyperdreamboothhypernetworksfastpersonalization,avrahami2023bas,wang2024instantidzeroshotidentitypreservinggeneration,guo2024pulidpurelightningid,shah2026ziplorasubjectstyleeffectively}. DreamBooth \cite{ruiz2023dreambooth} and Textual Inversion \cite{gal2022image} bind visual concepts to learned text tokens via per-subject fine-tuning. ControlNet \cite{zhang2023adding} further enables geometry-aware editing through auxiliary structural conditioning. Break-A-Scene~\cite{avrahami2023bas} also employs token optimization to decompose scene into tokens based on the masks and can achieve identity-preserving editing. 

Our approach performs 3D editing via 2D multi-view image editing. We apply textual inversion and DreamBooth fine-tuning to learn object-specific concepts, then generate edited multi-view images using a sparse-view diffusion model conditioned on learned text embeddings.

\subsection{Text-Guided 3D Editing}

The text-guided 3D editing has been explored widely in the research community including gaussian-based, mesh-based ones as well as the ones leveraging 2D diffusion priors ~\cite{wang2024gaussianeditorediting3dgaussians,michel2021text2meshtextdrivenneuralstylization,mikaeili2023skedsketchguidedtextbased3d,haque2023instructnerf2nerfediting3dscenes, wu2024gaussctrl, zheng2025pro3deditorprogressiveviewsperspective,ye2025nano3d,mvedit2024,barda2024instant3ditmultiviewinpaintingfast,chen2024partgenpartlevel3dgeneration}. Vox-E \cite{sella2023vox} performs volumetric editing by distilling diffusion guidance into a voxel grid, enabling localized modifications through a volumetric attention mechanism. However, its reliance on score distillation sampling (SDS) makes optimization prohibitively slow, requiring approximately one hour per edit. MVEdit \cite{mvedit2024} proposes a multi-view editing pipeline that leverages 2D diffusion models to edit rendered views and reconstructs the result into 3D. While MVEdit produces strong texture edits, it struggles to preserve object identity when changing the geometry of the object, such as pose modifications and redesign, often producing unrealistic colors.

To achieve text-guided fast 3D editing, PrEditor3D \cite{erkocc2025preditor3d} applies DDPM inversion and Prompt-to-Prompt within a sparse multi-view diffusion model, followed by feed-forward 3D reconstruction via GTR \cite{zhuang2024gtr} trained on large-scale datasets such as Objaverse. 

While efficient, it offers limited editing capability for keeping the shape consistent. In contrast, our approach learns object-level token embeddings for each object, enabling object-level editing (\eg, reshaping or pose changes) and supports color/texture changes. Furthermore, our method is substantially faster than optimization-based approaches: training requires approximately a few minutes, and once trained, each edit including textured mesh reconstruction takes roughly two minutes at inference time.

\subsection{Textual Inversion and Personalization}

Textual inversion \cite{gal2022image} introduced the idea of learning new token embeddings to represent specific visual concepts within the vocabulary of a pre-trained text-to-image diffusion model. The approach has been investigated further down the line~\cite{voynov2023p+,kumari2022customdiffusion,ruiz2023dreambooth}. DreamBooth \cite{ruiz2023dreambooth} extends this idea by fine-tuning the diffusion model itself with a class-specific prior preservation loss, achieving higher fidelity to the target concept. Custom Diffusion \cite{kumari2022customdiffusion} further improves efficiency by fine-tuning only the cross-attention layers and supports composing multiple concepts. Break-A-Scene \cite{avrahami2023bas} takes a complementary approach by learning \emph{multiple} disentangled token embeddings from a single image, where each token captures a distinct object or region guided by segmentation masks and an attention-based loss that encourages spatial correspondence between tokens and image regions. In the 3D domain, DreamBooth3D \cite{raj2023dreambooth3d} adapts DreamBooth for 3D-consistent generation from multi-view images. Break-A-Scene's disentangled token learning operates in the 2D domain. Our work extends this paradigm to 3D editing by operating on multi-view images rendered from a 3D mesh and introducing joint multi-view training on MVDream to ensure 3D-consistent token representations.

\subsection{Multi-View Diffusion Models}
Multi-view diffusion models generate consistent images from multiple viewpoints, enabling 3D-aware generation~\cite{liu2023zero, liu2023syncdreamer, shi2023mvdream, wang2023imagedreamimagepromptmultiviewdiffusion,wonder3dpp,long2023wonder3dsingleimage3d,gao2024cat3dcreate3dmultiview}. Zero-1-to-3 \cite{liu2023zero} fine-tunes a diffusion model to synthesize novel views given a single input image and a relative camera transformation. SyncDreamer \cite{liu2023syncdreamer} generates synchronized multi-view images by modeling cross-view attention within the diffusion process. MVDream \cite{shi2023mvdream} extends text-to-image diffusion models to generate multi-view consistent images by finetuning on large-scale 3D data, achieving strong 3D coherence while retaining the generative power of the 2D prior. We adopt MVDream \cite{shi2023mvdream} as our multi-view diffusion backbone and introduce a joint multi-view training strategy that optimizes disentangled token embeddings across all views simultaneously, ensuring that the learned representations maintain 3D consistency.

\section{Method}
\label{sec:method}

\begin{figure*}[t]
  \centering
  \includegraphics[width=\textwidth]{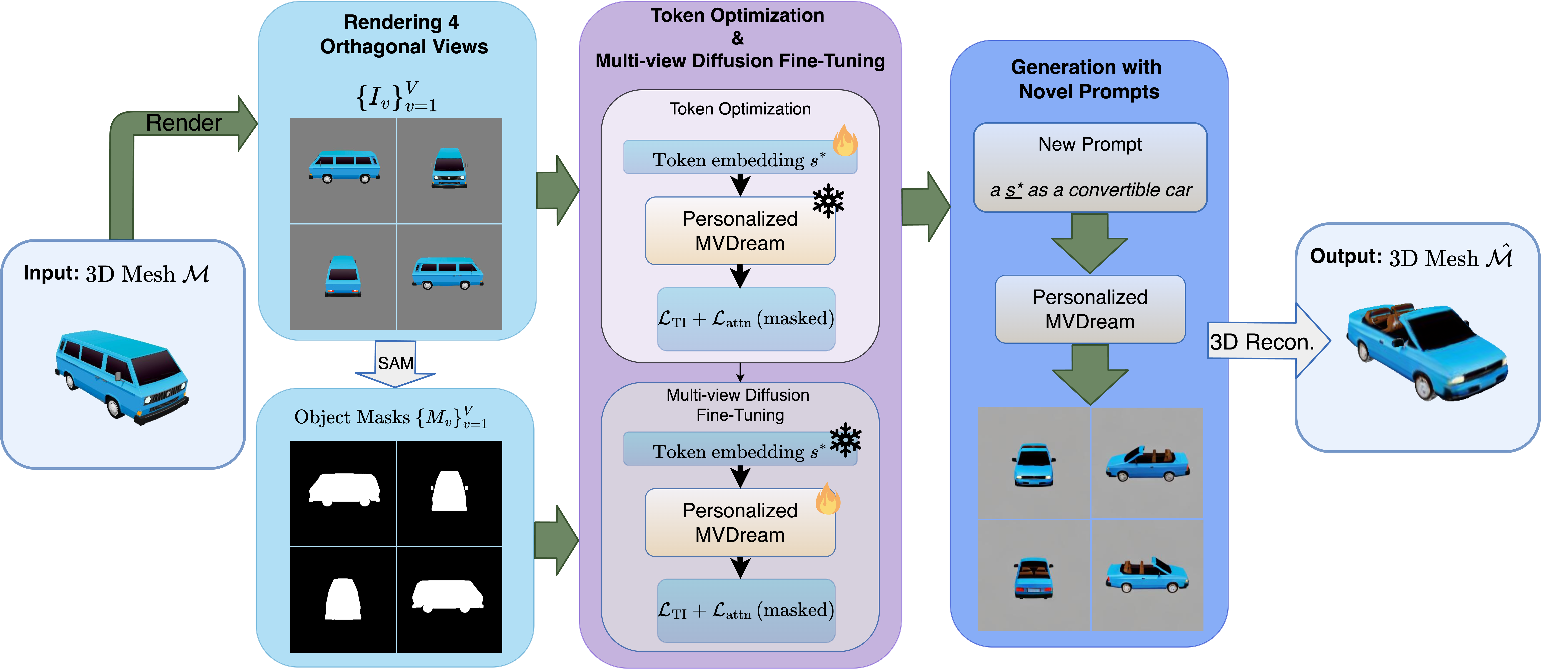}
  \caption{\textbf{Method overview.} \textbf{Top:} Given a 3D mesh, we render four orthogonal views and obtain object masks via SAM. \textbf{Middle:} In Phase~1 (TI), a token embedding $s^*$ is learned for the object through textual inversion on 4 views with a frozen UNet and attention alignment loss. In Phase~2 (DB), the full UNet is fine-tuned jointly across all 4 views with prior preservation. \textbf{Bottom:} At inference, tokens are composed with edit prompts, MVDream generates 4 consistent edited views, and GTR reconstructs the final 3D mesh.}
  \label{fig:method}
\end{figure*}

Given a 3D object represented as a textured mesh, our goal is to enable compositional, object-level editing through natural language. Our method consists of four stages: (1) rendering multi-view images and obtaining object semantic masks from the input mesh, (2a) learning disentangled token embeddings for each object via a two-phase optimization, (2b) jointly fine-tuning across views in Phase~2 to ensure 3D consistency, and (3) composing the learned tokens with editing prompts to generate modified consistent multi-view images that are reconstructed into an edited 3D mesh. An overview of our method is shown in Fig.~\ref{fig:method}.

\subsection{Multi-View Rendering}
\label{sec:rendering}
Given an input 3D mesh $\mathcal{M}$, we render a set of $N{=}4$ views $\{I_v\}_{v=1}^{N}$ at orthogonal azimuths starting from $90^\circ$ with a span of $360^\circ$ and a fixed elevation of $15^\circ$, using a differentiable renderer. This four-view configuration aligns with the camera setup used by MVDream \cite{shi2023mvdream}, which was trained to generate four consistent views.

\subsection{Token Optimization}
\label{sec:token-learning}

Our token optimization extends Break-A-Scene \cite{avrahami2023bas} to 3D by operating within MVDream~\cite{shi2023mvdream} and introducing joint cross-view training for 3D consistency. We learn a token embedding $s^*$ initialized from a semantically related word (\eg, ``robot,'' ``dog'') and optimize it in two phases.

\paragraph{Textual Inversion.}
In the first phase, we optimize only the token embedding $s^*$ while keeping the entire UNet frozen. This phase operates on four orthogonal view images to efficiently learn the initial token representation. We optimize $s^*$ by minimizing a masked diffusion denoising objective:
\begin{equation}
\mathcal{L}_{\text{TI}} = \mathbb{E}_{v, \epsilon, t} \left[ \| (\epsilon - \epsilon_\theta(z_t^{v}, t, c(y))) \odot m \|_2^2 \right] + \mu \mathcal{L}_{\text{attn}},
\label{eq:textual_inversion}
\end{equation}
where $z_t^{v}$ is the noisy latent of the object image $I_v$ at diffusion timestep $t$, $\epsilon$ is the sampled noise, $\epsilon_\theta$ is the frozen UNet, $c(y)$ is the text conditioning with prompt $y = $ ``a photo of $s^*$'', and $m$ is the object mask downsampled to the latent resolution. The mask ensures that the loss focuses on the object region.

As proposed by \cite{avrahami2023bas}, the attention alignment loss $\mathcal{L}_{\text{attn}}$ encourages the cross-attention map of the learned token to spatially align with the ground-truth object mask:
\begin{equation}
\mathcal{L}_{\text{attn}} = \| A - \hat{M} \|_2^2,
\label{eq:attn_loss}
\end{equation}
where $A$ is the aggregated cross-attention map for token $s^*$ across UNet layers and $\hat{M}$ is the normalized ground-truth mask. This loss is weighted by $\mu$ and is applied only during Phase~1.

To prevent catastrophic drift of the pre-trained vocabulary, we restore the embeddings of all non-learnable tokens after each optimization step, ensuring that only the new token embedding is modified.

\paragraph{UNet Fine-Tuning.}
Textual inversion alone may not capture fine-grained geometric and textural details, as the expressiveness is limited to the token embedding space. In the second phase, we unfreeze the full UNet and continue to optimize both the UNet parameters $\theta$ and the token embedding jointly. We optimize using the masked denoising objective with prior preservation:
\begin{equation}
\mathcal{L}_{\text{FT}} = \mathbb{E}_{v, \epsilon, t} \left[ \| (\epsilon - \epsilon_{\theta'}(z_t^{v}, t, c(y))) \odot m \|_2^2 \right] + \lambda \mathcal{L}_{\text{prior}},
\label{eq:unet_finetuning}
\end{equation}
where $\theta'$ denotes the updated UNet parameters. The prior preservation loss $\mathcal{L}_{\text{prior}}$ \cite{ruiz2023dreambooth} regularizes the fine-tuning to prevent language drift:
\begin{equation}
\mathcal{L}_{\text{prior}} = \mathbb{E}_{\epsilon, t} \left[ \| \epsilon - \epsilon_{\theta'}(z_t^{\text{pr}}, t, c(y_{\text{pr}})) \|_2^2 \right],
\label{eq:prior}
\end{equation}
where $z_t^{\text{pr}}$ are latents from class-prior images generated by the frozen model and $y_{\text{pr}}$ is the class prompt. The attention alignment loss is not applied in Phase~2, as the UNet weights are now being modified.

Rather than processing each view independently, which can lead to view-dependent inconsistencies, we introduce a joint multi-view training strategy that leverages MVDream's multi-view generation architecture to enforce 3D consistency.

MVDream processes four views jointly through a shared UNet with cross-view attention layers that exchange information across viewpoints. We exploit this by constructing training batches that contain all four views of the object simultaneously. The four-view latents $\{z_t^{v}\}_{v=1}^{4}$ are passed through the UNet together with their corresponding camera embeddings $\{e_v\}_{v=1}^{4}$, allowing the cross-view attention to enforce consistency:
\begin{equation}
\mathcal{L}_{\text{MV}} = \mathbb{E}_{\epsilon, t} \left[ \sum_{v=1}^{4} \| \epsilon_v - \epsilon_{\theta'}(\{z_t^{v}\}_{v=1}^{4}, t, c(y), \{e_v\}_{v=1}^{4})_v \|_2^2 \right],
\label{eq:multiview}
\end{equation}
where $\epsilon_{\theta'}(\cdot)_v$ denotes the predicted noise for view $v$ from the joint forward pass.

\subsection{Compositional Editing and Reconstruction}
\label{sec:editing}

\paragraph{Prompt Composition.}
Given an editing instruction for an object, we construct a prompt that uses the learned tokens with the desired modification. For example, to redesign a sofa to be ``single seat,'' we compose the prompt: ``a photo of $s^*$ redesigned to single seat,'' where $s^*$ is the learned token for the object whose identity should be preserved. This allows fine-grained control over appearance, geometry, and pose.

\paragraph{Multi-View Generation.}
Using the prompt $y_{\text{edit}}$ and the fine-tuned UNet $\epsilon_{\theta'}$, we generate four consistent edited views $\{\hat{I}_v\}_{v=1}^{4}$ via the standard MVDream sampling procedure with classifier-free guidance \cite{cfg}:
\begin{equation}
\tilde{\epsilon}_v = \epsilon_{\theta'}(\{z_t^v\}, t, \varnothing, \{e_v\}) + w \cdot \left(\epsilon_{\theta'}(\{z_t^v\}, t, c(y_{\text{edit}}), \{e_v\}) - \epsilon_{\theta'}(\{z_t^v\}, t, \varnothing, \{e_v\})\right),
\label{eq:cfg}
\end{equation}
where $w$ is the guidance scale and $\varnothing$ denotes the null text condition.

\paragraph{3D Reconstruction.}
The four generated views are reconstructed into a textured 3D mesh using GTR \cite{zhuang2024gtr}, a transformer-based large reconstruction model. GTR encodes the multi-view images into a triplane representation \cite{chan2022efficient} via a transformer generator, extracts geometry through differentiable marching cubes \cite{shen2021deep}, and applies a lightweight per-instance texture refinement procedure.

\section{Experiments}
\label{sec:experiments}

\subsection{Experimental Setup}
\label{sec:setup}

\paragraph{Benchmark.}
We construct a benchmark of 25 diverse editing cases spanning different object categories and edit types. Each case consists of a source 3D mesh paired with a text editing prompt. The benchmark covers a range of editing scenarios including attribute transfer (``dog as a cat,'' ``dog as a pig''), style transfer (``koala in lego style''), pose modifications (``cheetah lying on the floor,'' ``robot sitting''), object addition (``basket with apples,'' ``cake in a plate''), appearance editing (``person smile with teeth,'' ``person wearing sunglasses''), and shape redesign (``sofa redesigned to single seat'').

\paragraph{Baselines.}
We compare against three state-of-the-art text-guided 3D editing methods: \textbf{MVEdit} \cite{mvedit2024}, which applies 2D diffusion edits to rendered multi-view images and reconstructs the result into 3D; \textbf{Vox-E} \cite{sella2023vox}, which performs volumetric editing by distilling diffusion guidance into a voxel grid representation; and \textbf{PrEditor3D} \cite{erkocc2025preditor3d}, a training-free method that leverages 3D priors for fast editing.

\paragraph{Evaluation Metrics.}
Following the evaluation protocol of PrEditor3D \cite{erkocc2025preditor3d}, we evaluate all methods using the following metrics, computed over 70 rendered views per object:
\begin{itemize}
  \item \textbf{CLIP Directional Similarity} (CLIP$_\text{dir}$): Measures whether the change from the original to the edited rendering aligns with the change from the source prompt to the edit prompt in CLIP embedding space \cite{stylegan_nada}. A positive value indicates that the edit moves in the correct semantic direction.
  \item \textbf{CLIP Directional Cosine} (CLIP$_\text{dir-cos}$): The cosine similarity variant of the directional metric, providing a normalized measure of edit alignment.
  \item \textbf{CLIP Directional Avg} (CLIP$_\text{dir-avg}$): First averages the per-view image direction across all rendered views, then computes the dot product with the text direction. This reduces noise from individual views.
  \item \textbf{CLIP Directional Avg-Cosine} (CLIP$_\text{dir-avg-cos}$): The cosine similarity variant of CLIP$_\text{dir-avg}$, providing a normalized measure of the averaged edit direction alignment.
  \item \textbf{GPT-4V}: VLM-based metric to evaluate the quality of the generated shapes from the renderings. We measure quality in terms of: Prompt Alignment, 3D Plausibility, Identity Preservation, Visual Quality, 3D Consistency, Completeness, and Overall Quality.
\end{itemize}

\paragraph{Implementation Details.}
We use MVDream \cite{shi2023mvdream} (sd-v2.1-base-4view) as our multi-view diffusion backbone. In token optimisation, we optimize the token embeddings for 400 steps on single-view images with a learning rate of $5 \times 10^{-4}$, an attention alignment weight of $\mu = 10^{-2}$, and masked diffusion loss. In multi-view diffusion fine-tuning, we unfreeze the full UNet and jointly optimize it with the token embeddings for 400 steps with a learning rate of $2 \times 10^{-6}$, a prior preservation weight of $\lambda = 1.0$, and joint multi-view training across 4 views. We use the AdamW optimizer ($\beta_1{=}0.9$, $\beta_2{=}0.999$, weight decay $10^{-2}$) with 8-bit quantization and FP16 mixed precision. All images are rendered at $256 \times 256$ resolution with camera elevation $15^\circ$, azimuth starting at $90^\circ$, and $360^\circ$ span. For multi-view generation, we use a classifier-free guidance scale of $w = 7.5$ and 50 DDIM sampling steps \cite{song2020denoising}. For 3D reconstruction, we use GTR \cite{zhuang2024gtr} with a triplane resolution of $32{\times}32$ with 40 channels, marching cubes at $256^3$ resolution, and texture refinement loss weights $\alpha{=}0.5$, $\gamma{=}1.0$, $\delta{=}0.2$, $\eta{=}0.5$.

\subsection{Comparison with Baselines}
\label{sec:comparison}

\paragraph{Quantitative Results.}
Table~\ref{tab:benchmark} reports CLIP-based results averaged across all 25 benchmark cases. Our method achieves the highest CLIP directional similarity (3.16), CLIP$_\text{dir-cos}$ (11.84), and CLIP$_\text{dir-avg-cos}$ (15.98), indicating that our edits most faithfully follow the intended semantic direction. Vox-E is the second best on directional metrics but substantially lower, suggesting that while its outputs match the target text, the edits do not consistently move in the correct semantic direction. MVEdit and PrEditor3D achieve lower scores across all directional metrics, as they apply relatively conservative edits.

Table~\ref{tab:gpt4v} reports GPT-4V evaluation results. Our method achieves the highest scores across all seven dimensions, with particularly strong margins on Prompt Alignment (8.60 vs.\ 5.36 for the next best) and Completeness (8.46 vs.\ 7.48). MVEdit achieves the second-highest Visual Quality (6.56) and 3D Consistency (7.16), reflecting its conservative editing strategy that preserves appearance but limits edit expressiveness. Vox-E scores the lowest across most metrics, particularly on Visual Quality (3.92) and 3D Consistency (5.00), due to the artifacts introduced by its voxel-based optimization. PrEditor3D achieves the highest Identity Preservation (5.96) among baselines, but its overall quality remains lower than ours. These results confirm that our personalization approach enables more faithful and higher-quality edits while preserving the input object's identity.

\begin{table}[t]
  \centering
  \caption{Quantitative comparison averaged over all evaluation cases. Best results are in \textbf{bold}. Metrics are scaled by $\times 100$.}
  \label{tab:benchmark}
  \resizebox{0.65\linewidth}{!}{%
  \begin{tabular}{l c c c c}
    \toprule
    Method & CLIP$_\text{dir}$$\uparrow$ & CLIP$_\text{dir-cos}$$\uparrow$ & CLIP$_\text{dir-avg}$$\uparrow$ & CLIP$_\text{dir-avg-cos}$$\uparrow$ \\
    \midrule
    MVEdit \cite{mvedit2024}             & 1.77 & 6.57 & 1.77 & 8.42 \\
    Vox-E \cite{sella2023vox}            & 2.41 & 6.91 & 2.41 & 8.30 \\
    PrEditor3D \cite{erkocc2025preditor3d} & 1.26 & 5.05 & 1.26 & 7.26 \\
    Ours                                 & \textbf{3.16} & \textbf{11.84} & \textbf{3.16} & \textbf{15.98} \\
    \bottomrule
  \end{tabular}%
  }
\end{table}

\begin{table}[t]
  \centering
  \caption{GPT-4V evaluation averaged over all evaluation cases. Each method is scored out of 10, for the given metric. Best results are in \textbf{bold}.}
  \label{tab:gpt4v}
  \resizebox{\linewidth}{!}{%
  \begin{tabular}{l c c c c c c c}
    \toprule
    Method & Prompt Algn.$\uparrow$ & 3D Plaus.$\uparrow$ & Identity$\uparrow$ & Vis.\ Qual.$\uparrow$ & 3D Consist.$\uparrow$ & Complete.$\uparrow$ & Overall$\uparrow$ \\
    \midrule
    MVEdit \cite{mvedit2024}               & 5.36 & 6.84 & 5.60 & 6.56 & 7.16 & 7.48 & 6.32 \\
    Vox-E \cite{sella2023vox}              & 5.32 & 4.56 & 5.12 & 3.92 & 5.00 & 5.54 & 4.40 \\
    PrEditor3D \cite{erkocc2025preditor3d} & 4.84 & 5.68 & 5.96 & 5.29 & 6.04 & 6.83 & 5.36 \\
    Ours                                   & \textbf{8.60} & \textbf{7.28} & \textbf{7.67} & \textbf{6.62} & \textbf{7.54} & \textbf{8.46} & \textbf{7.40} \\
    \bottomrule
  \end{tabular}%
  }
\end{table}

\begin{table}[t]
  \centering
  \caption{User study results (30 participants). Each cell shows the percentage of times our method was preferred over the baseline. Higher is better.}
  \label{tab:userstudy}
  \scalebox{0.8}{

  \begin{tabular}{l c c c}
    \toprule
    vs Baseline & Prompt Algn.$\uparrow$ & Vis.\ Qual.$\uparrow$ & Shape Pres.$\uparrow$ \\
    \midrule
    MVEdit \cite{mvedit2024}               & 80.6\% & 75.5\% & 77.0\% \\
    PrEditor3D \cite{erkocc2025preditor3d} & 88.8\% & 82.4\% & 80.7\% \\
    Vox-E \cite{sella2023vox}              & 89.9\% & 94.5\% & 90.3\% \\
    \bottomrule
  \end{tabular}}
\end{table}

\paragraph{User Study.}
We conduct a user study to complement our automatic metrics. Our benchmark covers 15 objects across 25 editing cases, each paired with 3 baselines, yielding 75 pairwise comparisons in total. For each comparison, participants are asked three questions: prompt alignment, visual quality, and shape preservation. Each participant is randomly assigned 20 out of the 75 comparisons. We collected responses from 30 participants, resulting in 600 pairwise judgments. As shown in Table~\ref{tab:userstudy}, our method is consistently preferred over all baselines across all three dimensions. The margin is largest against Vox-E (89.9--94.5\%), whose voxel-based optimization frequently introduces visual artifacts. Against MVEdit, the preference is smallest on visual quality (75.5\%), consistent with its conservative editing strategy that preserves appearance at the cost of limited edit expressiveness. Against PrEditor3D, our method is strongly preferred on prompt alignment (88.8\%), reflecting PrEditor3D's difficulty in achieving the intended semantic changes.

\begin{figure*}[h!]
  \centering
  \includegraphics[width=\linewidth]{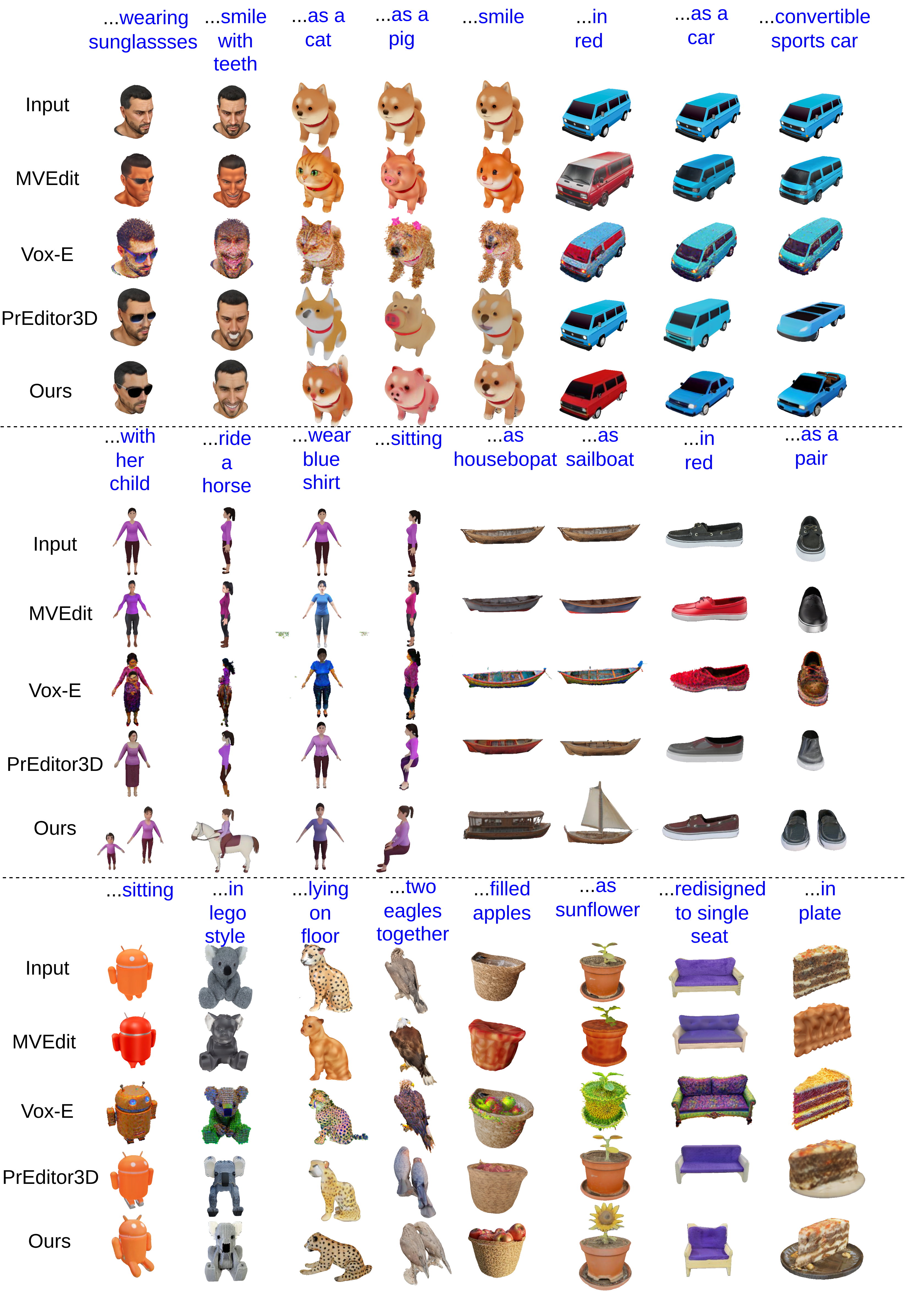}
  \caption{\textbf{Qualitative comparison.}}
  \label{fig:benchmark}
\end{figure*}

\paragraph{Qualitative Results.}
Figure~\ref{fig:benchmark} presents qualitative comparisons across representative benchmark cases. In the “van convertible” example, only our method successfully opens the roof, whereas all three baselines leave the van unchanged. For “two eagles together,” MVEdit and Vox-E fail to generate a second eagle, while PrEditor3D produces two instances but exhibits noticeable color drift. In “cake in a plate,” MVEdit and Vox-E omit the plate entirely, and although PrEditor3D introduces one, it appears blurry. In the “sofa single seat” case, only our method faithfully follows the prompt, correctly transforming the sofa into a single-seat design. For “person wearing sunglasses,” MVEdit and Vox-E generate unrealistic outputs, while PrEditor3D fails to produce sunglasses altogether. Across all scenarios, our method demonstrates superior adherence to the editing prompts while consistently preserving object identity.

To further highlight the versatility of our approach, we showcase a diverse set of editing tasks. These include attribute transfer (“dog as a cat,” “dog as a pig”), where species are altered while maintaining the original pose; object addition (“basket with apples”), where new elements are coherently integrated; pose modification (“robot sitting,” “lady sitting,” “lady riding a horse”), where body configurations are adjusted; appearance editing (“dog smiling,” “shoes in red”), targeting fine-grained attributes; style transfer (“koala in LEGO style”), altering material appearance; and structural transformations (“boat to houseboat,” “boat with sails,” “lady with child”), involving substantial geometric changes. In all cases, the learned token embedding effectively preserves the core identity of the input object while enabling precise and flexible modifications.

\subsection{Ablation Studies}
\label{sec:ablation}

\begin{table*}[t]
  \centering
  \caption{Ablation study on the \emph{Robot Sitting} case. Best results per group are in \textbf{bold}. $\uparrow$: higher is better. CLIP metrics are scaled by $\times 100$.}
  \label{tab:ablation}
  \resizebox{\textwidth}{!}{%
  \begin{tabular}{l c c c c c c c c c c c}
    \toprule
    & \multicolumn{4}{c}{CLIP} & \multicolumn{7}{c}{GPT-4V} \\
    \cmidrule(lr){2-5} \cmidrule(lr){6-12}
    Configuration & Dir$\uparrow$ & Dir-cos$\uparrow$ & Dir-avg$\uparrow$ & Dir-avg-cos$\uparrow$ & Pr.\ Algn.$\uparrow$ & 3D Pl.$\uparrow$ & Ident.$\uparrow$ & Vis.\ Q.$\uparrow$ & 3D Con.$\uparrow$ & Compl.$\uparrow$ & Overall$\uparrow$ \\
    \midrule
    \multicolumn{12}{l}{\textit{Multi-view joint training}} \\
    \quad Front view only       & $-$0.39 & $-$2.76 & $-$0.39 & $-$3.92 & 3 & 4 & 8 & 5 & 4 & 6 & 5 \\
    \quad Back view only        & $-$0.49 & $-$2.88 & $-$0.49 & $-$4.09 & 2 & 3 & 6 & 3 & 4 & 5 & 3 \\
    \quad Side view only        & $-$0.39 & $-$1.39 & $-$0.39 & $-$1.81 & 2 & 3 & 5 & 3 & 4 & 4 & 3 \\
    \quad 4 orthogonal views (Ours) & \textbf{0.51} & \textbf{3.15} & \textbf{0.51} & \textbf{4.79} & \textbf{8} & \textbf{7} & \textbf{9} & \textbf{6} & \textbf{8} & \textbf{9} & \textbf{7} \\
    \midrule
    \multicolumn{12}{l}{\textit{Masked losses}} \\
    \quad w/o cross attention loss   & 0.11 & 0.71 & 0.11 & 0.73 & 3 & 5 & 6 & 4 & 5 & 6 & 5 \\
    \quad w/o masked diffusion loss  & $-$0.27 & $-$2.43 & $-$0.27 & $-$3.38 & 2 & 3 & 7 & 4 & 5 & 6 & 4 \\
    \quad w/o both mask loss      & $-$0.41 & $-$2.15 & $-$0.41 & $-$2.72 & 3 & 5 & 7 & 4 & 6 & 7 & 5 \\
    \quad Full (Ours)          & \textbf{0.51} & \textbf{3.15} & \textbf{0.51} & \textbf{4.79} & \textbf{8} & \textbf{7} & \textbf{9} & \textbf{6} & \textbf{7} & \textbf{8} & \textbf{7} \\
    \midrule
    \multicolumn{12}{l}{\textit{Two-phase optimization}} \\
    \quad w/o DB (TI only, Phase~1)   & \textbf{4.80} & \textbf{15.77} & \textbf{4.80} & \textbf{17.77} & 2 & 3 & 1 & 3 & 4 & 5 & 3 \\
    \quad w/o TI (DB only, Phase~2)   & 0.40 & 2.53 & 0.40 & 3.80 & \textbf{9} & \textbf{7} & 8 & 6 & 7 & \textbf{8} & 7 \\
    \quad TI + DB (Ours)      & 0.51 & 3.15 & 0.51 & 4.79 & \textbf{9} & \textbf{7} & \textbf{9} & \textbf{7} & \textbf{8} & 8 & \textbf{8} \\
    \bottomrule
  \end{tabular}%
  }
\end{table*}

\begin{figure*}[t]
  \centering
  \includegraphics[width=\textwidth]{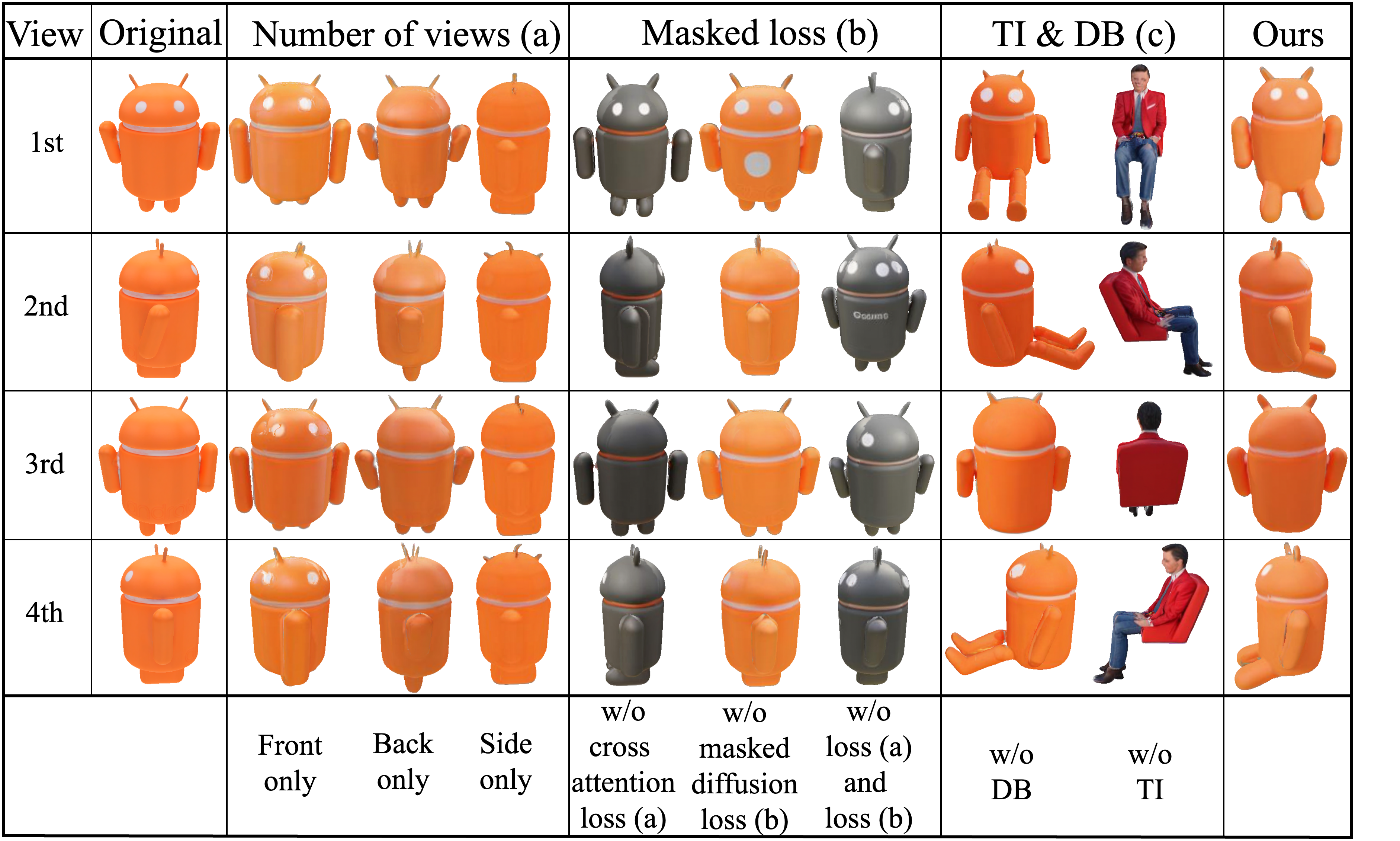}
  \caption{\textbf{Qualitative ablation study} on the \emph{Robot Sitting} case (``a photo of robot'' $\rightarrow$ ``a photo of robot sitting''). \textbf{(a)} Training with only a single view (front, back, or side) reduces 3D consistency. \textbf{(b)} Removing mask-based losses degrades edit localization. \textbf{(c)} Without TI, identity is partially lost; TI-only (no DreamBooth) fails to preserve the object.}
  \label{fig:ablation_visual}
\end{figure*}

We conduct a comprehensive ablation study to assess the contribution of each component. All experiments are performed on the \emph{Robot Sitting} case (a photo of a robot'' $\rightarrow$ a photo of a robot sitting''). Table~\ref{tab:ablation} presents the qualitative results, while Figure~\ref{fig:ablation_visual} provides corresponding visual comparisons.

\paragraph{Ablation (a): Multi-view training.}
As shown in Fig.~\ref{fig:ablation_visual}, when multi-view training is removed and only a single view is used, the edits become inconsistent across views due to the Janus problem. Specifically, when only the front view is used for fine-tuning, the white dots (eyes) from the front appear on all four generated views. Conversely, when only the back view is used, the generated views lose the white dots entirely. When only a single side view is used, every generated view exhibits arms, replicating the side-view features across all viewpoints. This demonstrates that single-view fine-tuning, regardless of which view is chosen, introduces the Janus problem by propagating view-specific features to all viewpoints. We also tested jointly fine-tuning with 2 and 3 views: as more views are jointly used, the Janus problem diminishes, yielding increasingly consistent results across viewpoints.

\paragraph{Ablation (b): Masked losses.}
As shown in Fig.~\ref{fig:ablation_visual}, each masking component addresses a distinct failure mode. Without the cross-attention loss, the token embedding absorbs background statistics, causing the reconstructed robot to shift toward the gray background color. Without the masked diffusion loss, the denoising objective is no longer restricted to the foreground region, leading to identity degradation manifested as spurious artifacts (\eg, a white dot on the robot's abdomen). When both losses are disabled, the two failure modes compound: the output suffers from both background color bleeding and structural artifacts, demonstrating the complementary roles of these losses.

\paragraph{Ablation (c): Two-phase optimization.}
As shown in Fig.~\ref{fig:ablation_visual}, the two optimization phases serve complementary roles. Without textual inversion (DB only), the token embedding is not optimized to represent the input object, so the model lacks a learned concept of the robot. As a result, the generated shape deviates from the original structure , the robot's legs become human-like, as the model defaults to its generic prior for ``sitting.'' Without DreamBooth fine-tuning (TI only), the multi-view diffusion model is not adapted to the learned token, and the strong semantic prior of ``sitting'' dominates: the model generates a generic sitting human, entirely discarding the robot's identity.

Notably, as shown in Table~\ref{tab:ablation}, the TI-only configuration achieves substantially higher CLIP directional scores (CLIP$_\text{dir}$: 4.80 vs.\ 0.51 for ours). However, this is misleading: without DreamBooth, the model produces a human in a canonical sitting pose, which strongly aligns with the ``sitting'' keyword in CLIP space despite completely failing to preserve the input object. The GPT-4V evaluation reveals this discrepancy TI only scores the lowest on Identity Preservation (1/10) and Prompt Alignment (2/10), confirming that high CLIP directional scores do not necessarily reflect faithful editing when object identity is lost. Our combined two-phase pipeline achieves the best balance: token optimisation stage anchors the token to the input object's appearance, while multiview diffusion model fine-tuning stage adapts the generative model to faithfully edit the learned concept without sacrificing identity. These results demonstrate that all components are essential for achieving our final performance.

\paragraph{Limitations.}
Our approach has several limitations. First, MVDream operates at a fixed resolution of $256{\times}256$, which limits the quality of the generated multi-view images and consequently affects the fidelity of the reconstructed mesh. Second, our method struggles with highly complex scenes; editing within large-scale environments with many interacting objects remains challenging, though we believe this can be addressed through hierarchical decomposition in future work. Finally, our method relies on four orthogonal views at a single elevation, which is the default configuration of MVDream. Since GTR benefits from a larger number of input views at varying elevations, this constraint limits the reconstruction quality and could be alleviated by adopting a multi-view backbone that supports more flexible camera configurations.

\section{Conclusion}
\label{sec:conclusion}
We presented \ours{}, a disentangled personalization framework for text-guided 3D editing that learns object-specific token embeddings within a multi-view diffusion model. Our two-phase optimization textual inversion followed by joint multi-view UNet fine-tuning on MVDream which enables diverse editing through natural language while preserving object identity. Quantitative evaluation on a benchmark of 25 diverse editing cases demonstrates that our method achieves the highest CLIP directional similarity and GPT-4V scores compared to MVEdit, Vox-E, and PrEditor3D, indicating stronger semantic alignment with the target edits. A user study with 30 participants further confirms that our method is consistently preferred over all baselines in prompt alignment, visual quality, and shape preservation. Ablation studies validate the importance of each component: joint multi-view training ensures 3D consistency, masked losses prevent background artifacts, and the two-phase pipeline balances edit fidelity with identity preservation.

\paragraph{Future Work.}
Several promising directions remain. First, incorporating 3D-aware attention mechanisms directly into the token learning phase and adopting stronger reconstruction backbones could improve the final mesh quality. Second, fine-tuning MVDream to support super-resolution output or integrating novel-view synthesis models such as Zero-1-to-3 \cite{liu2023zero} would increase the resolution and diversity of generated views. Third, inspired by PartCraft~\cite{ng2024partcraftcraftingcreativeobjects}, recombining disentangled tokens across different objects could enable creative cross-object composition, while leveraging attribute-level control over the learned tokens would allow more precise specification of color and texture. Finally, applying a zoom-in strategy to focus on local regions could extend our framework to handle large-scale scene editing.

\section*{Acknowledgements}
We thank Prof.\ Matthias Nie{\ss}ner for his support and for providing the research environment at the Visual Computing Lab, and Ziya Erko\c{c} for his supervision, guidance, and valuable feedback throughout this project.

\bibliographystyle{splncs04}
\bibliography{main}

\begin{thebibliography}{10}
\providecommand{\url}[1]{\texttt{#1}}
\providecommand{\urlprefix}{URL }
\providecommand{\doi}[1]{https://doi.org/#1}

\bibitem{avrahami2023bas}
Avrahami, O., Aberman, K., Fried, O., Cohen-Or, D., Lischinski, D.:
  Break-a-scene: Extracting multiple concepts from a single image. In: SIGGRAPH
  Asia 2023 Conference Papers. SA '23, Association for Computing Machinery, New
  York, NY, USA (2023). \doi{10.1145/3610548.3618154},
  \url{https://doi.org/10.1145/3610548.3618154}

\bibitem{baldridge2024imagen}
Baldridge, J., Bauer, J., Bhutani, M., Brichtova, N., Bunner, A., Castrejon,
  L., Chan, K., Chen, Y., Dieleman, S., Du, Y., et~al.: Imagen 3. arXiv
  preprint arXiv:2408.07009  (2024)

\bibitem{barda2024instant3ditmultiviewinpaintingfast}
Barda, A., Gadelha, M., Kim, V.G., Aigerman, N., Bermano, A.H., Groueix, T.:
  Instant3dit: Multiview inpainting for fast editing of 3d objects (2024),
  \url{https://arxiv.org/abs/2412.00518}

\bibitem{BetkerImprovingIG}
Betker, J., Goh, G., Jing, L., TimBrooks, Wang, J., Li, L., LongOuyang,
  JuntangZhuang, JoyceLee, YufeiGuo, WesamManassra, PrafullaDhariwal, CaseyChu,
  YunxinJiao, Ramesh, A.: Improving image generation with better captions.
  \url{https://api.semanticscholar.org/CorpusID:264403242}

\bibitem{chan2022efficient}
Chan, E.R., Lin, C.Z., Chan, M.A., Nagano, K., Pan, B., De~Mello, S., Gallo,
  O., Guibas, L.J., Tremblay, J., Khamis, S., et~al.: Efficient geometry-aware
  3d generative adversarial networks. In: Proceedings of the IEEE/CVF
  conference on computer vision and pattern recognition. pp. 16123--16133
  (2022)

\bibitem{mvedit2024}
Chen, H., Shi, R., Liu, Y., Shen, B., Gu, J., Wetzstein, G., Su, H., Guibas,
  L.: Generic 3d diffusion adapter using controlled multi-view editing (2024)

\bibitem{chen2024partgenpartlevel3dgeneration}
Chen, M., Shapovalov, R., Laina, I., Monnier, T., Wang, J., Novotny, D.,
  Vedaldi, A.: Partgen: Part-level 3d generation and reconstruction with
  multi-view diffusion models (2024), \url{https://arxiv.org/abs/2412.18608}

\bibitem{erkocc2025preditor3d}
Erko{\c{c}}, Z., G{\"u}meli, C., Wang, C., Nie{\ss}ner, M., Dai, A., Wonka, P.,
  Lee, H.Y., Zhuang, P.: Preditor3d: Fast and precise 3d shape editing. In:
  Proceedings of the IEEE/CVF Conference on Computer Vision and Pattern
  Recognition. pp. 640--649 (2025)

\bibitem{esser2024scalingrectifiedflowtransformers}
Esser, P., Kulal, S., Blattmann, A., Entezari, R., Müller, J., Saini, H.,
  Levi, Y., Lorenz, D., Sauer, A., Boesel, F., Podell, D., Dockhorn, T.,
  English, Z., Lacey, K., Goodwin, A., Marek, Y., Rombach, R.: Scaling
  rectified flow transformers for high-resolution image synthesis (2024),
  \url{https://arxiv.org/abs/2403.03206}

\bibitem{gal2022image}
Gal, R., Alaluf, Y., Atzmon, Y., Patashnik, O., Bermano, A.H., Chechik, G.,
  Cohen-Or, D.: An image is worth one word: Personalizing text-to-image
  generation using textual inversion. arXiv preprint arXiv:2208.01618  (2022)

\bibitem{stylegan_nada}
Gal, R., Patashnik, O., Maron, H., Bermano, A.H., Chechik, G., Cohen-Or, D.:
  {StyleGAN-NADA}: {CLIP}-guided domain adaptation of image generators. In: ACM
  Transactions on Graphics (TOG). vol.~41, pp. 1--13 (2022)

\bibitem{gao2024cat3dcreate3dmultiview}
Gao, R., Holynski, A., Henzler, P., Brussee, A., Martin-Brualla, R.,
  Srinivasan, P., Barron, J.T., Poole, B.: Cat3d: Create anything in 3d with
  multi-view diffusion models (2024), \url{https://arxiv.org/abs/2405.10314}

\bibitem{goodfellow2020generative}
Goodfellow, I., Pouget-Abadie, J., Mirza, M., Xu, B., Warde-Farley, D., Ozair,
  S., Courville, A., Bengio, Y.: Generative adversarial networks.
  Communications of the ACM  \textbf{63}(11),  139--144 (2020)

\bibitem{guo2024pulidpurelightningid}
Guo, Z., Wu, Y., Chen, Z., Chen, L., Zhang, P., He, Q.: Pulid: Pure and
  lightning id customization via contrastive alignment (2024),
  \url{https://arxiv.org/abs/2404.16022}

\bibitem{haque2023instructnerf2nerfediting3dscenes}
Haque, A., Tancik, M., Efros, A.A., Holynski, A., Kanazawa, A.:
  Instruct-nerf2nerf: Editing 3d scenes with instructions (2023),
  \url{https://arxiv.org/abs/2303.12789}

\bibitem{hertz2022prompt}
Hertz, A., Mokady, R., Tenenbaum, J., Aberman, K., Pritch, Y., Cohen-Or, D.:
  Prompt-to-prompt image editing with cross attention control. arXiv preprint
  arXiv:2208.01626  (2022)

\bibitem{cfg}
Ho, J., Salimans, T.: Classifier-free diffusion guidance. arXiv preprint
  arXiv:2207.12598  (2022)

\bibitem{hong2023lrm}
Hong, Y., Zhang, K., Gu, J., Bi, S., Zhou, Y., Liu, D., Liu, F., Sunkavalli,
  K., Bui, T., Tan, H.: Lrm: Large reconstruction model for single image to 3d.
  arXiv preprint arXiv:2311.04400  (2023)

\bibitem{karras2019style}
Karras, T., Laine, S., Aila, T.: A style-based generator architecture for
  generative adversarial networks. In: Proceedings of the IEEE/CVF conference
  on computer vision and pattern recognition. pp. 4401--4410 (2019)

\bibitem{kirillov2023segany}
Kirillov, A., Mintun, E., Ravi, N., Mao, H., Rolland, C., Gustafson, L., Xiao,
  T., Whitehead, S., Berg, A.C., Lo, W.Y., Doll{\'a}r, P., Girshick, R.:
  Segment anything. arXiv:2304.02643  (2023)

\bibitem{kumari2022customdiffusion}
Kumari, N., Zhang, B., Zhang, R., Shechtman, E., Zhu, J.Y.: Multi-concept
  customization of text-to-image diffusion  (2023)

\bibitem{lin2023magic3d}
Lin, C.H., Gao, J., Tang, L., Takikawa, T., Zeng, X., Huang, X., Kreis, K.,
  Fidler, S., Liu, M.Y., Lin, T.Y.: Magic3d: High-resolution text-to-3d content
  creation. In: Proceedings of the IEEE/CVF conference on computer vision and
  pattern recognition. pp. 300--309 (2023)

\bibitem{liu2023zero}
Liu, R., Wu, R., Van~Hoorick, B., Tokmakov, P., Zakharov, S., Vondrick, C.:
  Zero-1-to-3: Zero-shot one image to 3d object. In: Proceedings of the
  IEEE/CVF international conference on computer vision. pp. 9298--9309 (2023)

\bibitem{liu2023syncdreamer}
Liu, Y., Lin, C., Zeng, Z., Long, X., Liu, L., Komura, T., Wang, W.:
  Syncdreamer: Generating multiview-consistent images from a single-view image.
  arXiv preprint arXiv:2309.03453  (2023)

\bibitem{long2023wonder3dsingleimage3d}
Long, X., Guo, Y.C., Lin, C., Liu, Y., Dou, Z., Liu, L., Ma, Y., Zhang, S.H.,
  Habermann, M., Theobalt, C., Wang, W.: Wonder3d: Single image to 3d using
  cross-domain diffusion (2023), \url{https://arxiv.org/abs/2310.15008}

\bibitem{lugmayr2022repaint}
Lugmayr, A., Danelljan, M., Romero, A., Yu, F., Timofte, R., Van~Gool, L.:
  {RePaint}: Inpainting using denoising diffusion probabilistic models. In:
  Proceedings of the IEEE/CVF Conference on Computer Vision and Pattern
  Recognition. pp. 11461--11471 (2022)

\bibitem{michel2021text2meshtextdrivenneuralstylization}
Michel, O., Bar-On, R., Liu, R., Benaim, S., Hanocka, R.: Text2mesh:
  Text-driven neural stylization for meshes (2021),
  \url{https://arxiv.org/abs/2112.03221}

\bibitem{mikaeili2023skedsketchguidedtextbased3d}
Mikaeili, A., Perel, O., Safaee, M., Cohen-Or, D., Mahdavi-Amiri, A.: Sked:
  Sketch-guided text-based 3d editing (2023),
  \url{https://arxiv.org/abs/2303.10735}

\bibitem{ng2024partcraftcraftingcreativeobjects}
Ng, K.W., Zhu, X., Song, Y.Z., Xiang, T.: Partcraft: Crafting creative objects
  by parts (2024), \url{https://arxiv.org/abs/2407.04604}

\bibitem{poole2022dreamfusion}
Poole, B., Jain, A., Barron, J.T., Mildenhall, B.: Dreamfusion: Text-to-3d
  using 2d diffusion. arXiv preprint arXiv:2209.14988  (2022)

\bibitem{raj2023dreambooth3d}
Raj, A., Kaza, S., Poole, B., Niemeyer, M., Ruiz, N., Mildenhall, B., Zada, S.,
  Aberman, K., Rubinstein, M., Barron, J., et~al.: Dreambooth3d: Subject-driven
  text-to-3d generation. In: Proceedings of the IEEE/CVF international
  conference on computer vision. pp. 2349--2359 (2023)

\bibitem{ramesh2022hierarchical}
Ramesh, A., Dhariwal, P., Nichol, A., Chu, C., Chen, M.: Hierarchical
  text-conditional image generation with {CLIP} latents. In: arXiv preprint
  arXiv:2204.06125 (2022)

\bibitem{rombach2022latent}
Rombach, R., Blattmann, A., Lorenz, D., Esser, P., Ommer, B.: High-resolution
  image synthesis with latent diffusion models. In: Proceedings of the IEEE/CVF
  Conference on Computer Vision and Pattern Recognition. pp. 10684--10695
  (2022)

\bibitem{ruiz2023dreambooth}
Ruiz, N., Li, Y., Jampani, V., Pritch, Y., Rubinstein, M., Aberman, K.:
  Dreambooth: Fine tuning text-to-image diffusion models for subject-driven
  generation. In: Proceedings of the IEEE/CVF conference on computer vision and
  pattern recognition. pp. 22500--22510 (2023)

\bibitem{ruiz2024hyperdreamboothhypernetworksfastpersonalization}
Ruiz, N., Li, Y., Jampani, V., Wei, W., Hou, T., Pritch, Y., Wadhwa, N.,
  Rubinstein, M., Aberman, K.: Hyperdreambooth: Hypernetworks for fast
  personalization of text-to-image models (2024),
  \url{https://arxiv.org/abs/2307.06949}

\bibitem{sella2023vox}
Sella, E., Fiebelman, G., Hedman, P., Averbuch-Elor, H.: Vox-e: Text-guided
  voxel editing of 3d objects. In: Proceedings of the IEEE/CVF international
  conference on computer vision. pp. 430--440 (2023)

\bibitem{shah2026ziplorasubjectstyleeffectively}
Shah, V., Ruiz, N., Cole, F., Lu, E., Lazebnik, S., Li, Y., Jampani, V.:
  Ziplora: Any subject in any style by effectively merging loras (2026),
  \url{https://arxiv.org/abs/2311.13600}

\bibitem{shen2021deep}
Shen, T., Gao, J., Yin, K., Liu, M.Y., Fidler, S.: Deep marching tetrahedra: a
  hybrid representation for high-resolution 3d shape synthesis. Advances in
  Neural Information Processing Systems  \textbf{34},  6087--6101 (2021)

\bibitem{shi2023mvdream}
Shi, Y., Wang, P., Ye, J., Long, M., Li, K., Yang, X.: Mvdream: Multi-view
  diffusion for 3d generation. arXiv preprint arXiv:2308.16512  (2023)

\bibitem{song2020denoising}
Song, J., Meng, C., Ermon, S.: Denoising diffusion implicit models. arXiv
  preprint arXiv:2010.02502  (2020)

\bibitem{voynov2023p+}
Voynov, A., Chu, Q., Cohen-Or, D., Aberman, K.: p+: Extended textual
  conditioning in text-to-image generation. arXiv preprint arXiv:2303.09522
  (2023)

\bibitem{wang2024gaussianeditorediting3dgaussians}
Wang, J., Fang, J., Zhang, X., Xie, L., Tian, Q.: Gaussianeditor: Editing 3d
  gaussians delicately with text instructions (2024),
  \url{https://arxiv.org/abs/2311.16037}

\bibitem{wang2023imagedreamimagepromptmultiviewdiffusion}
Wang, P., Shi, Y.: Imagedream: Image-prompt multi-view diffusion for 3d
  generation (2023), \url{https://arxiv.org/abs/2312.02201}

\bibitem{wang2024instantidzeroshotidentitypreservinggeneration}
Wang, Q., Bai, X., Wang, H., Qin, Z., Chen, A., Li, H., Tang, X., Hu, Y.:
  Instantid: Zero-shot identity-preserving generation in seconds (2024),
  \url{https://arxiv.org/abs/2401.07519}

\bibitem{wang2023prolificdreamer}
Wang, Z., Lu, C., Wang, Y., Bao, F., Li, C., Su, H., Zhu, J.: Prolificdreamer:
  High-fidelity and diverse text-to-3d generation with variational score
  distillation. Advances in neural information processing systems  \textbf{36},
   8406--8441 (2023)

\bibitem{wu2024gaussctrl}
Wu, J., Bian, J.W., Li, X., Wang, G., Reid, I., Torr, P., Prisacariu, V.A.:
  Gaussctrl: Multi-view consistent text-driven 3d gaussian splatting editing.
  In: European conference on computer vision. pp. 55--71. Springer (2024)

\bibitem{wonder3dpp}
Yang, Y., Long, X.X., Dou, Z., Lin, C., Liu, Y., Yan, Q., Ma, Y., Wang, H., Wu,
  Z., Yin, W.: Wonder3d++: Cross-domain diffusion for high-fidelity 3d
  generation from a single image. IEEE Transactions on Pattern Analysis and
  Machine Intelligence  \textbf{48}(2),  1674–1688 (Feb 2026).
  \doi{10.1109/tpami.2025.3618675},
  \url{http://dx.doi.org/10.1109/TPAMI.2025.3618675}

\bibitem{ye2023ip}
Ye, H., Zhang, J., Liu, S., Han, X., Yang, W.: Ip-adapter: Text compatible
  image prompt adapter for text-to-image diffusion models. arXiv preprint
  arXiv:2308.06721  (2023)

\bibitem{ye2025nano3d}
Ye, J., Xie, S., Zhao, R., Wang, Z., Yan, H., Zu, W., Ma, L., Zhu, J.: Nano3d:
  A training-free approach for efficient 3d editing without masks. arXiv
  preprint arXiv:2510.15019  (2025)

\bibitem{zhang2023adding}
Zhang, L., Rao, A., Agrawala, M.: Adding conditional control to text-to-image
  diffusion models. In: Proceedings of the IEEE/CVF international conference on
  computer vision. pp. 3836--3847 (2023)

\bibitem{zheng2025pro3deditorprogressiveviewsperspective}
Zheng, Y., Huang, M., Chen, N., Mao, Z.: Pro3d-editor : A progressive-views
  perspective for consistent and precise 3d editing (2025),
  \url{https://arxiv.org/abs/2506.00512}

\bibitem{zhuang2024gtr}
Zhuang, P., Han, S., Wang, C., Siarohin, A., Zou, J., Vasilkovsky, M.,
  Shakhrai, V., Korolev, S., Tulyakov, S., Lee, H.Y.: Gtr: Improving large 3d
  reconstruction models through geometry and texture refinement. arXiv preprint
  arXiv:2406.05649  (2024)

\end{thebibliography}
\newpage
\section{Appendix}

We organize the supplementary material as follows: Sec.~\ref{sec:supp_quality_time} analyzes the editing quality vs.\ editing time trade-off; Sec.~\ref{sec:supp_impl} describes additional implementation details; Sec.~\ref{sec:supp_losses} gives expanded loss formulations; Sec.~\ref{sec:supp_benchmark} lists all benchmark editing cases; and Sec.~\ref{sec:supp_userstudy} details our user study protocol.

\subsection{Editing Quality vs.\ Editing Time}
\label{sec:supp_quality_time}

\begin{figure}[h!]
  \centering
  \includegraphics[width=0.5\linewidth]{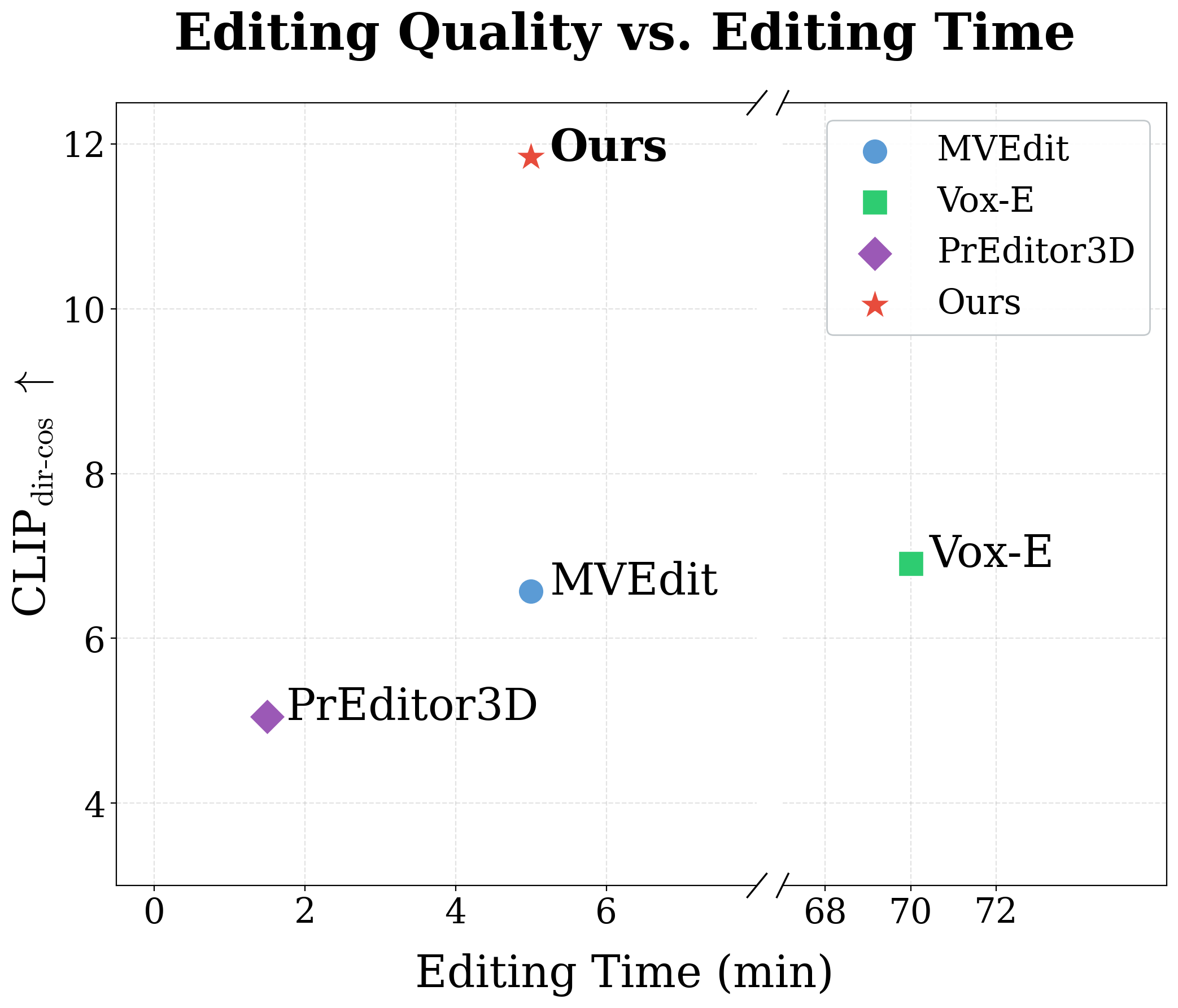}
  \caption{\textbf{Editing quality vs.\ editing time.} Our method achieves the highest CLIP$_{\text{dir-cos}}$ score while maintaining a competitive editing time of ${\sim}5$ minutes.}
  \label{fig:quality_vs_time}
\end{figure}

Figure~\ref{fig:quality_vs_time} illustrates the trade-off between editing quality (measured by CLIP$_{\text{dir-cos}}$) and computational cost across all compared methods. Our method achieves the highest editing fidelity while requiring only ${\sim}5$ minutes per edit comparable to MVEdit \cite{mvedit2024} and over an order of magnitude faster than Vox-E \cite{sella2023vox}, which demands ${\sim}70$ minutes due to its iterative SDS-based voxel optimization. PrEditor3D \cite{erkocc2025preditor3d} is the fastest at ${\sim}1.5$ minutes but attains the lowest editing quality, indicating that its speed comes at the expense of semantic accuracy. Our method occupies the ideal position in this quality time space, delivering state-of-the-art results without prohibitive computational overhead.

\subsection{Additional Implementation Details}
\label{sec:supp_impl}

\paragraph{Hardware.}
All experiments are conducted on a single NVIDIA GeForce RTX 3090 GPU (24\,GB VRAM) with CUDA 12.4.

\paragraph{Segmentation Pipeline.}
We use the Segment Anything Model (SAM)~\cite{kirillov2023segany} to obtain object-level masks from the rendered multi-view images. For each rendered view, we provide a single point prompt at the center of the object bounding box. The resulting masks are binarized at a threshold of 0.5 and downsampled to the latent resolution ($32{\times}32$) for use in the masked diffusion losses. No additional post-processing (\eg, morphological operations or CRF refinement) is applied, as SAM produces sufficiently clean masks for our use case.

\paragraph{Initializer Tokens.}
For each editing case, we initialize the learnable token embedding $s^*$ from a semantically related word in the pre-trained vocabulary. The initializer token is chosen to roughly match the object category (\eg, ``robot'' for the Robot Sitting case, ``dog'' for Dog as Cat/Pig, ``van'' for all van cases, ``person'' for sunglasses/smile cases). This provides a meaningful starting point for textual inversion and accelerates convergence compared to random initialization. The full list of initializer tokens for all 25 cases is provided in Table~\ref{tab:cases}.

\paragraph{Inference Prompts.}
During multi-view generation, we use the composed editing prompt as the positive condition (\eg, ``a photo of $s^*$ redesigned to single seat'') with a classifier-free guidance scale of $w{=}7.5$. All 25 editing prompts are listed in Table~\ref{tab:cases}.

\paragraph{GTR Reconstruction Details.}
For 3D reconstruction, we use GTR~\cite{zhuang2024gtr} with the following configuration: a triplane resolution of $32{\times}32$ with 40 feature channels, geometry extraction via differentiable marching cubes at $256^3$ resolution, and a per-instance texture refinement step. The texture refinement optimizes the triplane color features on the extracted mesh surface using the four generated views with a combination of losses: photometric RGB loss, perceptual LPIPS loss ($\alpha{=}0.5$), silhouette mask loss ($\gamma{=}1.0$), depth consistency loss ($\delta{=}0.2$), and opacity regularization ($\eta{=}0.5$). The reconstruction takes approximately one minute per object on the single NVIDIA GeForce RTX 3090.

\paragraph{Evaluation Rendering Setup.}
For quantitative evaluation, we render 70 views of each reconstructed mesh following the protocol of PrEditor3D~\cite{erkocc2025preditor3d}. The camera trajectory uniformly samples azimuths over a full $360^\circ$ rotation. All rendered images are at $256{\times}256$ resolution. CLIP-based metrics are computed on all 70 views and averaged.

\subsection{Detailed Loss Formulations}
\label{sec:supp_losses}
\subsubsection{Phase 1: Textual Inversion with Masked Losses}

In Phase~1, we freeze the UNet $\epsilon_\theta$ and optimize only the token embedding $s^*$. The overall Phase~1 objective combines a masked diffusion loss with a cross attention regularizer:
\begin{equation}
\mathcal{L}_{\text{TI}} = \mathbb{E}_{v, \epsilon, t} \left[ \| (\epsilon - \epsilon_\theta(z_t^{v}, t, c(y))) \odot m_v \|_2^2 \right] + \mu \, \mathcal{L}_{\text{attn}},
\label{eq:supp_ti}
\end{equation}
where $v \in \{1, \ldots, 4\}$ indexes the orthogonal views, $\epsilon \sim \mathcal{N}(0, \mathbf{I})$ is sampled noise, $t \sim \mathcal{U}(1, T)$ is a uniformly sampled diffusion timestep, $z_t^{v} = \alpha_t z_0^{v} + \sigma_t \epsilon$ is the noised latent of view $v$ following the DDPM noise schedule, $c(y)$ is the CLIP text encoding of the prompt $y = \text{``a photo of } s^*\text{''}$, and $m_v$ is the binary object mask for view $v$ downsampled to the latent resolution ($32{\times}32$). The element-wise product $\odot\, m_v$ restricts the loss to the foreground object region, preventing the token from encoding background information.

\paragraph{Cross Attention Loss.}
The cross attention loss~\cite{avrahami2023bas} encourages the learned token's cross-attention activation pattern to match the ground-truth object segmentation mask:
\begin{equation}
\mathcal{L}_{\text{attn}} = \| A_{s^*} - \hat{M} \|_2^2,
\label{eq:supp_attn}
\end{equation}
where $A_{s^*} \in [0,1]^{h \times w}$ is the cross-attention map for token $s^*$, aggregated by averaging across all UNet cross-attention layers and attention heads, and $\hat{M}$ is the ground-truth mask normalized to $[0,1]$. This loss is weighted by $\mu = 10^{-2}$ and applied \emph{only} during Phase~1. It is disabled in Phase~2 because the UNet weights are being modified, which would create conflicting gradients between the cross attention loss and the masked diffusion loss.

\paragraph{Vocabulary Preservation.}
To prevent catastrophic drift of the pre-trained vocabulary during token optimization, we restore the embeddings of all non-learnable tokens after each gradient step. Concretely, let $\mathbf{E} \in \mathbb{R}^{V \times d}$ be the full token embedding matrix with vocabulary size $V$ and embedding dimension $d$. After each optimizer step, we set $\mathbf{E}[i] \leftarrow \mathbf{E}_0[i]$ for all $i \neq i_{s^*}$, where $\mathbf{E}_0$ is the original pre-trained embedding matrix and $i_{s^*}$ is the index of the newly added token.

\subsubsection{Phase 2: Multi-View UNet Fine-Tuning}

In Phase~2, we unfreeze the full UNet and jointly optimize the UNet parameters $\theta$ and the token embedding $s^*$. The Phase~2 objective combines the masked diffusion loss with a prior preservation term:
\begin{equation}
\mathcal{L}_{\text{FT}} = \mathbb{E}_{v, \epsilon, t} \left[ \| (\epsilon - \epsilon_{\theta'}(z_t^{v}, t, c(y))) \odot m_v \|_2^2 \right] + \lambda \, \mathcal{L}_{\text{prior}},
\label{eq:supp_ft}
\end{equation}
where $\theta'$ denotes the updated UNet parameters. The prior preservation loss~\cite{ruiz2023dreambooth} regularizes the fine-tuning to prevent language drift and mode collapse:
\begin{equation}
\mathcal{L}_{\text{prior}} = \mathbb{E}_{\epsilon, t} \left[ \| \epsilon - \epsilon_{\theta'}(z_t^{\text{pr}}, t, c(y_{\text{pr}})) \|_2^2 \right],
\label{eq:supp_prior}
\end{equation}
where $z_t^{\text{pr}}$ are noisy latents of class-prior images generated by the \emph{frozen} pre-fine-tuning model, and $y_{\text{pr}}$ is the class prompt (\eg, ``a photo of a robot''). The prior loss is computed over the full image without masking, as it regularizes the model's general generation ability. We set $\lambda = 1.0$.

\paragraph{Joint Multi-View Training.}
Rather than processing each view independently, we construct training batches containing all four views simultaneously. This leverages MVDream's cross-view attention layers~\cite{shi2023mvdream} to enforce 3D consistency across viewpoints:
\begin{equation}
\mathcal{L}_{\text{MV}} = \mathbb{E}_{\epsilon, t} \left[ \sum_{v=1}^{4} \| \epsilon_v - \epsilon_{\theta'}(\{z_t^{v}\}_{v=1}^{4}, t, c(y), \{e_v\}_{v=1}^{4})_v \|_2^2 \right],
\label{eq:supp_mv}
\end{equation}
where $\{e_v\}_{v=1}^{4}$ are the camera pose embeddings for each viewpoint (azimuth $\in \{90^\circ, 180^\circ, 270^\circ, 360^\circ\}$, elevation $= 15^\circ$), and $\epsilon_{\theta'}(\cdot)_v$ denotes the predicted noise for view $v$ from the joint forward pass through the shared UNet. The cross-view attention mechanism in MVDream allows each view to attend to features from all other views, encouraging geometric consistency across the generated viewpoints.

\subsubsection{Inference: Classifier-Free Guidance}

At inference time, we use the fine-tuned UNet $\epsilon_{\theta'}$ with classifier-free guidance (CFG)~\cite{cfg} to generate edited multi-view images from the composed editing prompt $y_{\text{edit}}$:
\begin{equation}
\tilde{\epsilon}_v = \epsilon_{\theta'}(\{z_t^v\}, t, \varnothing, \{e_v\}) + w \cdot \left(\epsilon_{\theta'}(\{z_t^v\}, t, c(y_{\text{edit}}), \{e_v\}) - \epsilon_{\theta'}(\{z_t^v\}, t, \varnothing, \{e_v\})\right),
\label{eq:supp_cfg}
\end{equation}
where $w = 7.5$ is the guidance scale, $\varnothing$ is the null text condition, and $y_{\text{edit}}$ is the composed editing prompt (\eg, ``a photo of $s^*$ redesigned to single seat'').

\begin{table}[h!]
  \centering
  \caption{Hyperparameter for our method.}
  \label{tab:hyperparams}
  \scalebox{0.78}{
  \begin{tabular}{l l l}
    \toprule
    \textbf{Stage} & \textbf{Hyperparameter} & \textbf{Value} \\
    \midrule
    \multirow{5}{*}{Phase 1 (TI)} & Training steps & 400 \\
    & Learning rate & $5 \times 10^{-4}$ \\
    & Cross attention weight $\mu$ & $10^{-2}$ \\
    & Optimizer & AdamW ($\beta_1{=}0.9$, $\beta_2{=}0.999$) \\
    & Weight decay & $10^{-2}$ \\
    \midrule
    \multirow{5}{*}{Phase 2 (DB)} & Training steps & 400 \\
    & Learning rate & $2 \times 10^{-6}$ \\
    & Prior preservation weight $\lambda$ & 1.0 \\
    & Optimizer & AdamW (8-bit, FP16 mixed precision) \\
    & Number of training views & 4 (joint) \\
    \bottomrule
  \end{tabular}}
\end{table}

\subsection{Benchmark Cases}
\label{sec:supp_benchmark}

Table~\ref{tab:cases} lists all 25 editing cases in our benchmark with their source descriptions and edit prompts. The benchmark is designed to cover a diverse range of editing scenarios: attribute transfer (\eg, ``dog as a cat''), style transfer (``koala in lego style''), pose modification (``robot sitting,'' ``lady sitting''), object addition (``basket with apples,'' ``lady with child''), appearance editing (``shoes in red,'' ``van in red''), and structural redesign (``sofa redesigned to single seat,'' ``van as convertible sports car'').

\begin{table}[h!]
  \centering
  \caption{Benchmark editing cases with source descriptions and edit prompts.}
  \label{tab:cases}
  \resizebox{\linewidth}{!}{%
  \begin{tabular}{r l l l}
    \toprule
    \# & Case & Source Prompt & Edit Prompt \\
    \midrule
    1  & Basket Apples     & a photo of basket   & a photo of basket with apples in it \\
    2  & Cake in Plate     & a photo of cake     & a photo of cake in a plate \\
    3  & Cheetah Lying     & a photo of cheetah  & a photo of cheetah lying on the floor \\
    4  & Dog as Cat        & a photo of dog      & a photo of dog as a cat \\
    5  & Dog as Pig        & a photo of dog      & a photo of dog as a pig \\
    6  & Dog Smile         & a photo of dog      & a photo of dog smile \\
    7  & Eagle Two         & a photo of eagle    & a photo of two eagle together \\
    8  & Plant Sunflower   & a photo of plant    & a photo of plant as sunflower \\
    9  & Sofa Redesigned   & a photo of sofa     & a photo of sofa redesigned to single seat \\
    10 & House Snow        & a photo of house    & a photo of house covered with snow \\
    11 & Shoes Red         & a photo of shoes    & a photo of shoes in red \\
    12 & Shoes Pair        & a photo of shoes    & a photo of two shoes as a pair \\
    13 & Sunglasses        & a photo of person   & a photo of person wearing sunglasses \\
    14 & Person Smile      & a photo of person   & a photo of person smile with teeth \\
    15 & Robot Sitting     & a photo of a robot  & a photo of robot sitting \\
    16 & Boat Houseboat    & a photo of a boat   & a photo of boat as houseboat \\
    17 & Boat Sailboat     & a photo of a boat   & a photo of boat with sails \\
    18 & Van Red           & a photo of van      & a photo of van in red \\
    19 & Van as Car        & a photo of van      & a photo of van as a car \\
    20 & Van Convertible   & a photo of van      & a photo of van as convertible sports car \\
    21 & Lady with Child   & a photo of lady     & a photo of lady with her child \\
    22 & Lady Ride Horse   & a photo of lady     & a photo of lady ride a white horse \\
    23 & Lady Blue T-shirt & a photo of lady     & a photo of lady wearing blue T-shirt \\
    24 & Lady Sitting      & a photo of lady     & a photo of lady sitting \\
    25 & Koala Lego        & a photo of koala    & a photo of koala in lego style \\
    \bottomrule
  \end{tabular}%
  }
\end{table}

\newpage
\subsection{User Study Details}
\label{sec:supp_userstudy}

We provide additional details on the user study protocol summarized in the main paper.

\paragraph{Study Design.}
Our benchmark comprises 15 distinct source objects edited across 25 cases, each compared against 3 baselines (MVEdit, Vox-E, PrEditor3D), yielding 75 pairwise comparisons in total. For each comparison, participants view the original 3D object (rendered from a canonical viewpoint), followed by the editing results of both our method and one baseline, presented in randomized left/right order to avoid positional bias. The edit prompt is displayed alongside the renderings.

\paragraph{Evaluation Criteria.}
For each pairwise comparison, participants answer three forced-choice questions:
\begin{enumerate}
  \item \textbf{Prompt Alignment:} ``Which result better matches the editing instruction?'' Measures whether the edit faithfully achieves the intended modification described in the text prompt.
  \item \textbf{Visual Quality:} ``Which result looks more realistic and visually appealing?'' Assesses the overall rendering quality, including texture sharpness, color fidelity, and absence of artifacts.
  \item \textbf{Shape Preservation:} ``Which result better preserves the original object's shape and identity?''  Evaluates whether the core geometry and identity of the source object are retained after editing.
\end{enumerate}
For each question, participants select one of the two presented results or choose ``cannot decide,'' following a forced-choice protocol.

\paragraph{Participant Recruitment and Assignment.}
We recruited 30 participants with varying levels of familiarity with 3D content creation and computer graphics. Each participant was randomly assigned 20 out of the 75 pairwise comparisons, ensuring that each comparison was evaluated by at least 8 participants. The random assignment was stratified to ensure balanced coverage across all baselines and editing categories. This yielded a total of 600 pairwise judgments (30 participants $\times$ 20 comparisons) across 1,800 individual question responses (600 judgments $\times$ 3 questions).

\paragraph{Presentation Details.}
Renderings were shown at $256{\times}256$ resolution on a white background. For each comparison, we displayed four rendered views (front, right, back, left) of both the original object and the two competing editing results, allowing participants to assess 3D consistency. The study was conducted via an online interface, and participants were given no time limit per comparison. The average completion time was approximately 15 minutes per participant.

\end{document}